\documentclass[arxiv]{melba}
\usepackage{amsmath,amsfonts}
\providecommand{\Lcal}{\mathcal{L}}
\providecommand{\Dcal}{\mathcal{D}}

\providecommand{\rvy}{\mathbf{y}}
\providecommand{\rmG}{\mathbf{G}}
\providecommand{\rmB}{\mathbf{B}}
\providecommand{\rmX}{\mathbf{X}}
\providecommand{\rvc}{\mathbf{c}}

\providecommand{\R}{\mathbb{R}}

\melbaid{YYYY:NNN}  % This is provided upon by the publishing editor
\doi{10.59275/j.melba.2024-AAAA}
\melbaauthors{A. Keshavarzi}  % Note: this one is also used to set the pdf 'authors' metadata
\email{ali.keshavarzi@telecom-paris.fr}
\volume{3}
\firstpageno{1337}  % Communicated by the publishing editor
\melbayear{2026}  % The publication year
\datesubmitted{yyyy-m1-d1}  % Date submitted to MELBA: mm/yyyy
\datepublished{yyyy-m2-d2}  % Today's date: mm/yyyy

\ShortHeadings{BifDet}{A. Keshavarzi}

\title{BifDet: A 3D Bifurcation Detection Dataset for Airway-Tree Modeling}

\author{
	\firstname Ali \surname Keshavarzi\aff{1}\orcid{0000-1111-2222-3333}, \:
	\firstname Quentin \surname Bouniot\aff{1},
	\firstname Benjamin M. \surname Smith\aff{2},
	\firstname Elsa \surname Angelini\aff{1}
}
\affiliations{% <- trailing '%' to avoid unwanted indent
	\num 1 \addr  LTCI, Telecom Paris, 
  Insitut Polytechnique de Paris,
  Palaiseau, France \\
	\num 2 \addr  Department of Medicine,
  Columbia University,
  New York, USA \\
}

\abstract{Thoracic Computed Tomography (CT) scans offer detailed insights into the intricate branching network of the airway tree, which is essential for understanding various respiratory diseases. Airway bifurcations, where airway branches split, are crucial landmarks for understanding lung physiology, disease mechanisms and lesion localization. Despite the significance of bifurcation analysis, a notable lack of  datasets annotated for this task hinders the development of advanced automated specialized detection or segmentation tools. In this paper, we introduce BifDet, the first publicly-available dataset specialized for 3D airway bifurcation detection, filling a critical gap in existing resources. Our dataset comprises carefully annotated CT scans from the ATM22 open-access cohort with  bifurcation bounding boxes covering the parent and daughter branches. As a use-case for demonstrating the potential of BifDet, we fine-tune and evaluate RetinaNet and DETR for 3D airway bifurcations detection on CT scans. We provide detailed pipelines, including preprocessing steps and specific implementation design choices. Results are detailed over various categories of minimal bounding box sizes to serve as baseline to benchmark future research. BifDet annotations and source code for 3D bifurcation detection are shared on Zenodo and Github (TBC AFTER ACCEPTANCE)}

\keywords{Bifurcation Detection, Airway-tree modeling, Lung CT scan, Dataset}

\begin{document}

% top matter
\twocolumn[\maketitle]
% comment the preceedings and uncomment the following if the authors list + abstract is longer than one page
% \maketitle
% \twocolumn

\section{Introduction}
\label{sec:introduction}

\begin{figure*}[tbp]
\centering
\includegraphics[width=0.99\textwidth]{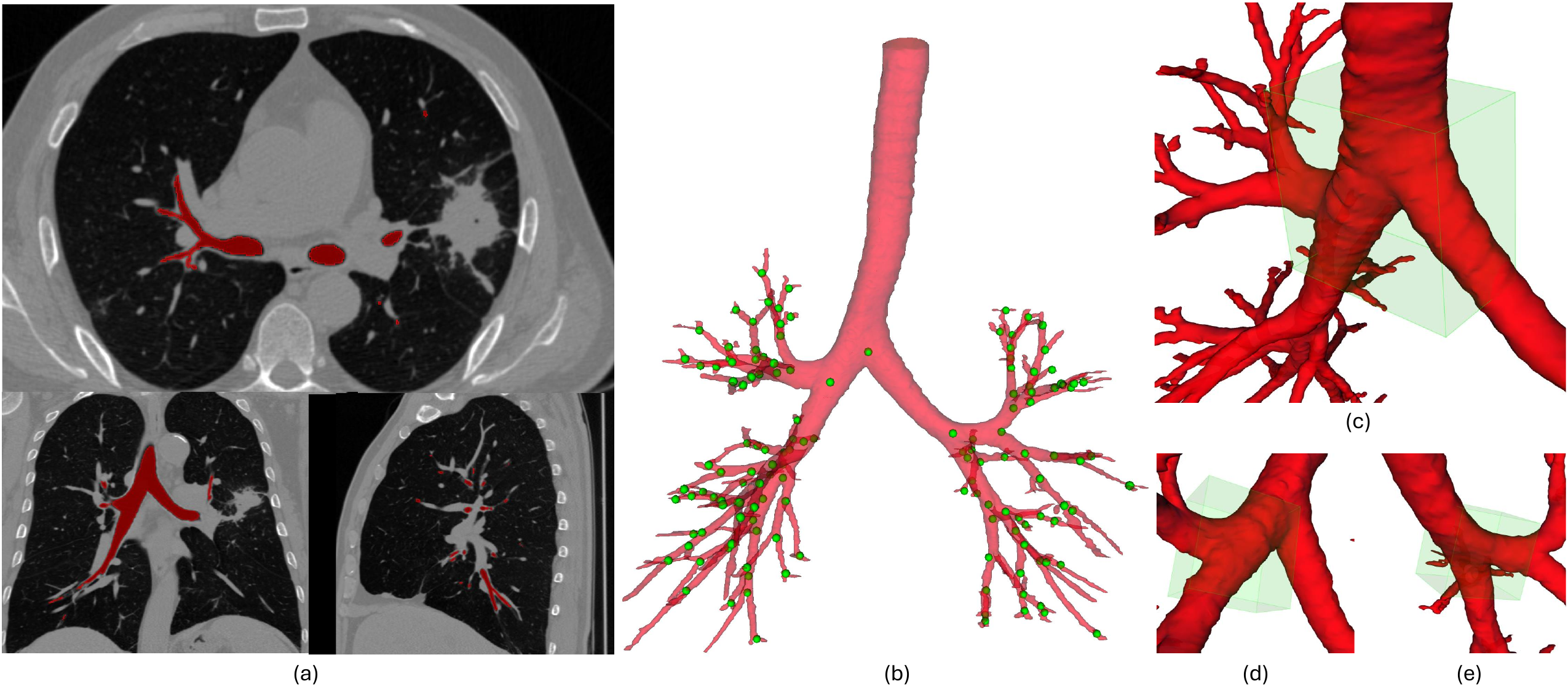}
\caption{(a) Lung CT scan with highlighted airway region. (b) 3D rendering of the airway tree with the centers of the ground-truth bounding boxes shown at bifurcation points. (c) Trachea ground-truth bifurcation bounding box. (d) and (e) Right and left main bronchi ground-truth bifurcation bounding boxes. 
% \textcolor{red}{should be changed: Copyright purposes}
}
\label{fig:dataset_intro}
\end{figure*}

Thoracic Computed Tomography (CT) scans provide detailed, high-resolution volumetric imaging of the lungs, offering clear views of their internal structures, including the airways.
As shown in Figure~\ref{fig:dataset_intro}, airways form a tree with tubular branches that vary in size and orientation, beginning with the major airways (trachea and two main bronchi) and progressively branching into smaller and distal airways. Clinicians describe and characterize the human airway tree into branch generations, based on the sequence of observed bifurcartions, starting from the trachea.

Precise modeling of the airway tree from CT scans is crucial for understanding lung disease pathophysiology and lung tumour localization~\citep{appmed_asthma2019, appmed_bronchiectasis2017, appmed_copd2018, appmed_emphysema2014, appmed_fibrosis2021, NEJMoa2414059}.

% airway features
Airway bifurcations, as illustrated in Figure~\ref{fig:dataset_intro}, correspond to locations where airway branches split.  Precise detection and localization of bifurcations is needed for airway tree labeling into generations \citep{airbif_wang2020} and quantification of branch-based lengths and diameters, and branching angles. These parameters enable measurements of biomarkers of clinical interest such as dysanapsis~\citep{medair_bokov2014}. 

Furthermore, 3D models of airway trees enable the simulation of airflow dynamics, gas exchange processes, and the distribution of inhaled substances within the lungs for studying respiratory physiology~\citep{cfd_kim2021,cfd_bass2021,cfd_Bertolini2024,cfd_KOLANJIYIL2017}, drug delivery~\citep{cfd_LONGEST2012,cfd_HOFMANN2011} and air pollutant susceptibility. 
They also guide bronchoscopy findings as in~\citep{bronchotrack} where patient-specific airway models are generated from pre-operative segmentation of airways in CT scans to construct a semantic airway graph for assigning branchial labels based on bifurcations~\citep{bronchotrack}. 

Few open-access datasets of lung CT scans  with manual annotation of airway trees exist~\citep{exact09,atm22,aiib23_ds, bas_dataset}. 
% \EAc{UPDATE HERE LIST OF ANNOTATED COHORTS: LIDC-IDRI}
These datasets have been used to train several segmentation methods~\citep{airseg_zhang2024,airseg_zheng2021,airseg_qin2019}.  
All methods still lack robustness and precision due to variability in CT image quality and limitations in supervised learning approaches for extracting topologically-complete and correct airway trees. 
Quality metrics used to evaluate airway segmentation methods focus on tree length and branch-level overlap. A natural quality check could focus on number of branching points detected or missed.
But automated extraction of branching points from ground-truth segmentations in publicly annotated CT lung datasets is cumbersome and error-prone.
Indeed these segmentations are not all topologically valid and vary greatly in segmentation depth in terms of number of branch generations. Direct exploitation of standard skeletonization algorithms ~\citep{LEE1994462,SAHA20163} fails to provide the correct number and precise location of branching points~\citep{graph_yu2023}. 

In this study, we annotated airway bifurcations on the ground-truth segmentations from two public datasets: EXACT09 and ATM22. Instead of using single points, we annotated bounding boxes covering the parent and daughter branches to capture the full morphological context and enable local measures of features of interest (eg. diameter and angles). 

% Contributions
We introduce BifDet, the first publicly-available annotation set dedicated to 3D airway bifurcations, addressing a significant gap in existing resources. We also propose a CT-based detection task formulation tailored for airway bifurcations, providing a standardized benchmark framework for future research on bifurcation detection. Our main contributions are summarized as follows:

\begin{itemize}
    \item \textbf{BifDet Dataset:} first publicly-available CT scan dataset annotated for 3D airway bifurcation detection.
    \item \textbf{Problem Formulation:} We propose a formal formulation of the detection task specifically tailored for airway bifurcations.
    \item \textbf{Methodological Pipeline:} We provide a comprehensive workflow for supervised training of airway bifurcation detection on CT scans, including image preprocessing, training, testing steps, and code implementations.
    \item \textbf{Benchmark Results:} We establish baseline detection performance across various detection metrics. 
\end{itemize}
We provide the annotations and the code on our GitHub repository for the automated  selection and setup of our training pipeline. 

\section{Related Works}
\subsection{Object Detection in Medical Imaging}

\begin{figure*}[tbp]
\centering
\includegraphics[width=0.99\textwidth]{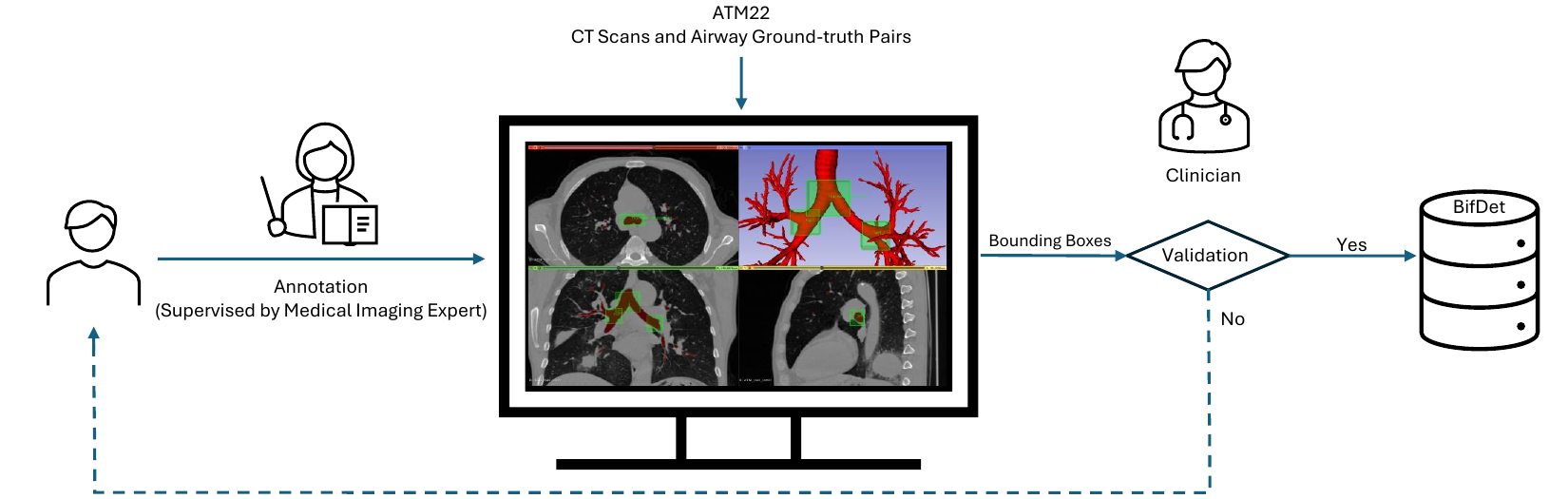}
\caption{Annotation pipeline. The average annotation time per case is approximately 8 hours (+2 hours of final check) depending on the complexity of the airway tree and the number of bifurcations.}
\label{fig:annot_pipeline}
\end{figure*}

Object detection models, including RetinaNet~\citep{retinanet_Lin2017}, Faster-RCNN~\citep{frcnn_ren2015}, DETR~\citep{detr_carion2020}, and Deformable DETR~\citep{defdetr_zhu2021}, have significantly improved medical image analysis, assisting the detection of lung nodules~\citep{medobjdetection_zhu2018, medobjdetection_xie2019}, lesions~\citep{medobjdetection_yan2018, medobjdetection_Zlocha2019, medobjdetection_shen2021}, and vascular abnormalities~\citep{medobjdetection_ma2021, medobjdetection_wu2020}. In CT-based applications, object detection has been extended to volumetric data, facilitating the transition from 2D to 3D bounding boxes~\citep{melba:2023:003:wittmann}. However, these approaches have not been widely explored for airway bifurcation detection.

\subsection{Vessel Bifurcation Detection} Bifurcation detection in medical imaging has been explored primarily in the context of blood vessels and the cardiovascular system~\citep{vessbif_rafic2023, vessbif_nouri2020, vessbif_zhao2014}. Techniques such as multi-scale filtering, Hessian-based methods~\citep{vessbif_shang2011}, and deep learning approaches have been developed to identify bifurcations with varying degrees of success. 
Zhao et al.~\citep{vessbif_zhao2017} tested their bifurcation detector, which operates by fitting a parametric geometric deformable model to 3D medical images, on both synthetic and private annotations built upon the Vessel12 challenge~\footnote{https://vessel12.grand-challenge.org} dataset. 
Wu et al.~\citep{vessbif_wu2016} introduced a generalized approach for vessel tracking and bifurcation detection, where a dedicated neural network classifies image regions as bifurcations, background, or vessel segments. The approach utilizes a vessel probability map and intensity profile analysis for bifurcation to determine the direction and center of subsequent tracking locations for each branch. 

\subsection{Airway Bifurcation Annotation \& Detection.} In the context of airways, however, bifurcation detection remains underexplored, with few non-specialized datasets and methodologies available that are limited to only the main branches (e.g. trachea, main right- and left-bronchi and first 2-3 generations)~\citep{airbif_wang2020}. 
SG-Net~\citep{sg_net2024} provides semantic segmentation annotations on EXACT09 (N=20 cases) and LIDC-IDRI (N=40 scans) with only 35 anatomical bifurcation landmark points up to the 4th branching generation. Furthermore, SG-Net assesses performance on initial branch generations using a private dataset that is not publicly available. 
Additionally, related works on airway bifurcation detection often involve estimating local branch directions which is challenging in small airways. 
A multi-loss deep learning network, TreeNet~\citep{airbif_zhao2018}, was introduced to predict branch directions and accurately extract 3D anatomical trees, such as airways, from medical images.
A subsequent extension of this work incorporated Long Short-Term Memory (LSTM) units into the model to better capture the sequential dependencies and long-term memory inherent in airway branching patterns, thus potentially improving bifurcation detection~\citep{airbif_zhao2019}.

Despite recent advances, existing airway bifurcation detection methods are constrained by limited annotated datasets, shallow branching depth coverage, and focus on semantic labeling or local direction estimation rather than direct bifurcation detection. These approaches are not easily reusable or generalizable. BifDet addresses this gap with large-scale, high-resolution 3D bounding-box annotations across full airway trees, enabling robust benchmarking and broader applicability.

\section{BifDet: Annotated Dataset}
\label{sec:bifdet_datasets}

\subsection{CT Data Source}
\label{subsec:data_aqui}

The BifDet dataset is derived from the ATM22~\footnote{https://atm22.grand-challenge.org} dataset. Each case in ATM22 includes ground-truth segmentation masks for the trachea, main bronchi, lobar bronchi, and distal segmental bronchi.
We randomly selected N=42 cases from the 300 publicly-available CT scans. 
Detailed information about case names, matrix dimension and voxel sizes, can be found in Appendix~\ref{appendix:bifdet_ATM22}. 
The CT scans and airway segmentation masks can be directly downloaded from the links provided in the original dataset paper~\citep{atm22}, . 
We refer the reader to the ATM22 paper and challenge website for further information regarding  data acquisition details. 
Sharing of the annotated dataset follows the copyright Creative Commons BY-NC-SA 4.0 license \copyright. 

\subsection{Bifurcation Annotation}
\label{subsec:bif_annot}

As shown in Figure~\ref{fig:annot_pipeline}, the manual annotation process is based on pairs of CT scans and ground-truth annotations visualised with a 3D rendering tool, namely 3D Slicer\footnote{\href{https://www.slicer.org}{https://www.slicer.org}}.
The manual annotation process for the BifDet dataset was meticulously carried out by a lung CT expert specializing in airway tree modeling, under the supervision of a medical imaging expert and a clinician specializing in pulmonary and respiratory health. 
Bounding boxes, aligned with image axis, were annotated to tightly encompass each airway bifurcation, adhering to the following criteria: (1) a bounding box covers the parent and all daughter branches; (2) a bounding box starts above where the bifurcation begins, i.e., where the parent starts diverging, and ends when the daughter branches are all identifiable; and (3) bounding boxes are allowed to overlap if necessary to ensure comprehensive coverage of the bifurcation structures.

\subsection{BifDet: Annotation Statistics}
\label{subsec:data_stat}

Our annotation dataset consists of N=7,523 bifurcations in total, with an average of 179 bifurcations per case (N=42) and with an average size of 6.39x6.23x6.21 mm or 7.96x7.76x12.43 voxels.

\section{Automated 3D Bifurcation Detection}
\label{sec:bifdet_task}
Compared to the pixel-level airway segmentation task, the branching bounding box detection task on BifDet presents a unique challenge due to the crowded nature of the scene, with a high number of bifurcations often clustered together, and the presence of extremely small objects, particularly in the distal regions of the airway trees. These factors necessitate the development of specialized detection algorithms capable of handling both crowded scenes and small object detection effectively.

\subsection{Notations}
We define the {3D bifurcation detection problem} as a localization and binary classification task.
Let $\mathcal{D} = \{(\rmX_i, \rvy_i, \rmG_i)\}_{i=1}^N$ represent our manually annotated dataset of $N$ CT scan volumes, where $\mathbf{X}_i \in \mathbb{R}^{W_i \times H_i \times D_i}$ is the \(i\)-th CT scan volume with matrix size $W_i$x$H_i$x$D_i$, and $\rvy_i = \{ 1 \}_{i=1}^{S_i}$ and $\rmG_i = \{\rmG_{(i,1)}, \rmG_{(i,2)}, \ldots, \rmG_{(i,S_i)}\}$ are the sets of ground-truth bounding box annotations, with $S_i$ the number of bifurcations in that volume.
Each bounding box is parameterized with a label $\rvy_i$ set to one if containing a bifurcation and a 6-D vector \( \rmG_{(i,j)} = (\rvc_{(i,j)}^*,w_{(i,j)}^*, h_{(i,j)}^*, d_{(i,j)}^*) \in \R^6 \) defined as: (1) its center coordinates \( \rvc_{(i,j)}^* = (x_{(i,j)}^*, y_{(i,j)}^*, z_{(i,j)}^*) \in \mathbb{R}^3 \), and (2) its dimensions in each direction (width, height, depth) \( (w_{(i,j)}^*, h_{(i,j)}^*, d_{(i,j)}^*) \in \mathbb{R}^3 \).

\subsection{Detection Methods}

\subsubsection{RetinaNet}
\label{subsubsec:retinanet}

As a first CNN-based baseline detector, we use RetinaNet~\citep{retinanet_Lin2017}, a simple one-stage detector based on Feature Pyramid Network (FPN)~\citep{lin2017feature} for feature extraction and two sub-networks for  classification and regression on input candidate boxes (anchors).

Initial anchors are set with various pre-defined sizes and aspect ratios, offering a diverse set of candidate for the network to evaluate.

The classification head predicts the likelihood of containing an airway bifurcation for each anchor, while the regression head predicts offsets to refine the anchor's location and size. 
% \textbf{Loss function.}
The learning objective is to minimize a linear combination of a classification loss $\Lcal_{BCE}$ based on binary cross-entropy for box labels, and a regression loss $\Lcal_{SL_1}$  based on Smooth $L_1$ metric  \citep{girshick2015fast} for the box size and position: 

\begin{equation}
\label{eq:retinanet_loss}
    \begin{aligned}
    \min_{\theta} \mathcal{L}(\Dcal) = \frac{1}{N} \sum_{i=1}^{N} \sum_{j=1}^{S_i} \frac{1}{S_i} [\lambda_{cls} \cdot \mathcal{L}_{BCE}(\hat{\rvy}_{(i,j)}, \rvy_{(i,j)}) \\ + \lambda_{reg} \cdot \mathcal{L}_{SL_1}(\rmB_{(i,j)}, \rmG_{(i,j)})],   
    \end{aligned}
\end{equation}
where $\lambda_{cls}$ and $\lambda_{reg}$ are hyperparameters used to balance the two loss terms.

\subsubsection{DETR}
\label{subsubsec:3detr}
As a second baseline detector, we use deformable DETR \citep{defdetr_zhu2021} which is a Transformer-based architecture extended for 3D detection in \citep{melba:2023:003:wittmann}. Transformer-based detectors, first introduced with DETR \citep{detr_carion2020}, are end-to-end object detection methods based on an encoder-decoder transformer architecture~\citep{vaswani2017attention} on top of a feature extractor.

Visual features are organized into a set of feature maps, which are then refined by a stack of Transformer encoders. Subsequently, a stack of Transformer decoders transforms a set of object queries that attend to the encoder output. Finally, separate classification and regression heads predict the class and bounding box position and size for each of these object queries. 
During training, Transformer-based detectors employ a bipartite matching process to associate the most accurate predicted box with the corresponding ground-truth labels. This process identifies the optimal permutation between predictions and ground-truths based on a weighted matching cost.
% \textbf{Loss function.} 
The overall general loss function combines a binary cross-entropy loss $\Lcal_{BCE}$ for the classification task, and two loss functions for the regression task, namely a scale-invariant Generalized IoU (GIoU) loss $\Lcal_{GIoU}$ and a $L_1$ error $\Lcal_{L_1}$: 

\begin{equation}
    \begin{aligned}
    \min_{\theta} \Lcal(\Dcal) = \frac{1}{N} \sum_{i=1}^{N} \sum_{j=1}^{S_i} \frac{1}{S_i} \biggl[ &\lambda_{cls} \cdot \Lcal_{BCE}(\hat{\rvy}_{(i,j)}, \rvy_{(i,j)}) \\
     & + \lambda_{GIoU} \cdot \Lcal_{GIoU}(\rmB_{(i,j)}, \rmG_{(i,j)})  \\ & + \lambda_{L1} \cdot \Lcal_{L_1}(\rmB_{(i,j)}, \rmG_{(i,j)}) \biggr],
    \end{aligned}
\end{equation}
where $\lambda_{cls}$, $\lambda_{GIoU}$ and $\lambda_{L1}$ are hyperparameters used to balance the loss terms.  

\section{Experiments}
\label{sec:experiments}

\subsection{Data preprocessing}
\label{subsubsec:preprocessing}

Pre-processing includes clipping of CT intensity values to the range of [-1000, 200], and cropping of the field of view around the lungs segmented using the tool from~\citep{lungsmaskhofmanninger2020}. We preserved the original isotropic axial voxel sizes (in the range of 0.74 to 0.90 mm) and the slice thickness of 0.5mm for all scans.
We randomly split the  N=42 annotated case into training  (N=33), and validation (N=9) subsets. Results are reported on the validation set.

\begin{table*}[tbh]
\footnotesize
\centering
\caption{Quantitative results for model performance measured by mAP, and AP/AR at different IoU thresholds (in \%, higher is better).}
\begin{tabular}{c|c|l|cccc|cccc}
\specialrule{0.2em}{1pt}{1pt} \addlinespace[4pt]
\textbf{Min. Box Dim} & \textbf{Avg. Remained Boxes}  & \textbf{Model} & \textbf{mAP} & $\textbf{AP}_{10}$ & $\textbf{AP}_{25}$ & $\textbf{AP}_{50}$ & \textbf{mAR} & $\textbf{AR}_{10}$ & $\textbf{AR}_{25}$ & $\textbf{AR}_{50}$ \\
\addlinespace[4pt] \specialrule{0.1em}{1pt}{1pt}

\multirow{2}{*}{4}  & \multirow{2}{*}{178.07}  & RetinaNet & \textbf{5.77}  & \underline{10.60} & 6.64  & 1.88  & \textbf{15.94} & \underline{23.90} & 18.59 & 7.51  \\
                    & & DETR      & 0.24  & 1.42  & 0.03  & 0.01  & 2.18  & 8.27  & 1.13  & 0.15  \\
\addlinespace[4pt] \specialrule{0.1em}{1pt}{1pt}

\multirow{2}{*}{6}  & \multirow{2}{*}{173.48}  & RetinaNet & \textbf{8.98}  & \underline{15.70} & 10.90 & 2.93  & \textbf{24.67} & \underline{35.58} & 28.95 & 12.50 \\
                    & & DETR      & 0.54  & 3.24  & 1.55  & 0.01  & 3.96  & 14.10 & 3.10  & 0.21  \\
\addlinespace[4pt] \specialrule{0.1em}{1pt}{1pt}

\multirow{2}{*}{8}  & \multirow{2}{*}{49.00}  & RetinaNet & \textbf{39.77} & \underline{54.86} & 45.84 & 17.67 & \textbf{54.68} & \underline{69.14} & 60.32 & 33.41 \\
                    & & DETR      & 2.20  & 8.26  & 1.73  & 0.02  & 9.87  & 28.77 & 9.05  & 1.16  \\
\addlinespace[4pt] \specialrule{0.1em}{1pt}{1pt}

\multirow{2}{*}{10} & \multirow{2}{*}{22.50}  & RetinaNet & \textbf{67.92} & \underline{74.66} & 73.08 & 47.96 & \textbf{76.93} & \underline{83.67} & 81.12 & 60.20 \\
                    & & DETR      & 2.35  & 9.57  & 1.37  & 0.01  & 15.14 & 43.37 & 15.31 & 1.02  \\

\specialrule{0.2em}{1pt}{1pt} \addlinespace[4pt]
\end{tabular}
\label{tbl:quant_res}
\end{table*}

\subsection{Implementation details}
\label{subsec:detection_models}

\subsubsection{3D Patch extraction}
During training, we extract random 3D patches of size $256 \times 256 \times 256$ voxels from each scan to introduce spatial variability and balance the distribution of bifurcation instances. 
At inference time, each validation scan is centrally cropped to a fixed 3D patch of size ($256 \times 256 \times 256$) voxels which captures the central and most critical region of the lungs for clinical biomarkers while including most bifurcations. 
On the validation scans used of this study, averaged number of ground-truth bifurcations with minimal box size threshold of 6 voxels is 114 versus 118 on full-lung FOV. At 10-voxel minimal size, we retain almost all GT bifurcations boxes.

The model outputs a set of predicted bounding boxes per scan, each associated with a confidence score reflecting the model's predicted confidence on the classification sub-task (i.e., the likelihood that a given box contains a bifurcation).

\subsubsection{RetinaNet}
\label{subsubsec:retinanet_impl}
The implementation of RetinaNet for airway bifurcation detection is based on a comprehensive framework provided by MONAI~\footnote{\url{https://monai.io/}}.
We set $\lambda_{cls}=0.5$ and $\lambda_{reg}=1$ to encourage the model to prioritize localization over classification. 

\textbf{Training and inference Setup.} 
RetinaNet uses a fixed set of predefined anchor boxes generated across multiple-scale feature maps. 
Specifically, we use 3-levels pyramid with spatial scales $[2^0, 2^1, 2^2]$, each associated with base anchor shapes of $(28\times19\times37$), ($23\times16\times24$), and ($20\times16\times20$) voxels, respectively. These anchors densely cover the input space to provide object proposals for bifurcation detection.
The model was optimized using the \textit{SGD} optimizer with a learning rate of \textit{10e-2}. A \textit{GradualWarmupScheduler} was applied for the first 10 epochs, followed by a \textit{StepLR} scheduler that reduced the learning rate by a factor of 10 every 50 epochs. To address class imbalance, we employed Adaptive Training Sample Selection (ATSS)~\citep{zhang2020bridging} with 4 candidate anchors per feature pyramid level. This was coupled with a hard negative sampler that selected 64 anchors per 3D patch, enforcing at least 16 negative anchors to guarantee stable background learning even when few positives were adaptively selected. 
% \EAc{why lower bound with negative and not positives}
Bounding box regression was performed directly in voxel space without normalization. During inference, non-maximum suppression (NMS) with a IoU threshold of \textit{0.22} was used to filter redundant predictions. The batch size was set to 1, and the model was trained for up to 200 epochs. To accelerate training and reduce memory usage, we utilized automatic mixed precision (AMP) and dataset caching. 
All experiments were conducted on a single NVIDIA A100 40GB GPU.

\subsubsection{DETR}
\label{subsubsec:3d_detr_impl}

We implemented a 3D adaptation of Deformable DETR for airway bifurcation detection. The architecture uses 3 encoder and 3 decoder layers, each with 6 attention heads and hidden dimension of 384. The transformer is fed with 300 learned object queries and is followed by feed-forward networks for classification and bounding box regression, each with 3 layers, hidden dimension 1024, and ReLU activations.

\textbf{Training and inference Setup.}:
Bounding box coordinates are normalized with respect to the image dimensions during training and rescaled back to voxel space during inference. This ensures stability during optimization and compatibility with COCO-style loss formulations.
The model was trained using the \textit{AdamW} optimizer~\citep{loshchilov2018decoupled}, with an initial learning rate of $2 \times 10^{-4}$ for the transformer and $2 \times 10^{-5}$ for the CNN backbone. The learning rate was reduced by a factor of 10 every 80 epochs. 
% \EAc{VERY SMALL VALUES. TO CONFIRM} 
Loss weights were set as $\lambda_{\text{cls}}=2$, $\lambda_{\text{GIoU}}=2$, and $\lambda_{\text{L1}}=5$.
To mitigate the dominance of background class predictions caused by the fixed number of queries, we down-weighted the ``no-object'' class in the loss function. Similar to RetinaNet, the model was trained for 200 epochs with a batch size of 1 using AMP, and all experiments were run on a single NVIDIA A100 40GB GPU.

\subsubsection{Evaluation metrics}
We employ well-established object detection metrics: Intersection over Union (IoU), mean Average Precision (mAP), and mean Average Recall (mAR). 
We follow the COCO evaluation protocol and use the official implementation from the COCO benchmark suite~\citep{lin2014microsoft}. Detailed definitions of these metrics are provided in Appendix~\ref{appendix:eval_metrics}.
Average Precision (AP) is computed for a given IoU threshold $T$ as the area under the precision-recall curve, evaluated over the model's confidence-ranked predictions. 
Average Recall (AR) is computed for a fixed parameter $N_D$, as the recall achieved when selecting the top-$N_D$ predictions based on confidence.
We report mAP and mAR as the mean of AP and AR across a range of IoU thresholds $T \in [0.1, 0.5]$ in increments of 0.05. The number of detections per image is fixed to $N_D = 300$ during inference for both models (compared to a maximum of 274 GT annotated bifurcations in training cases and 270 in validation cases).

\begin{figure*}[tbp]
\centering
\includegraphics[width=0.99\textwidth]{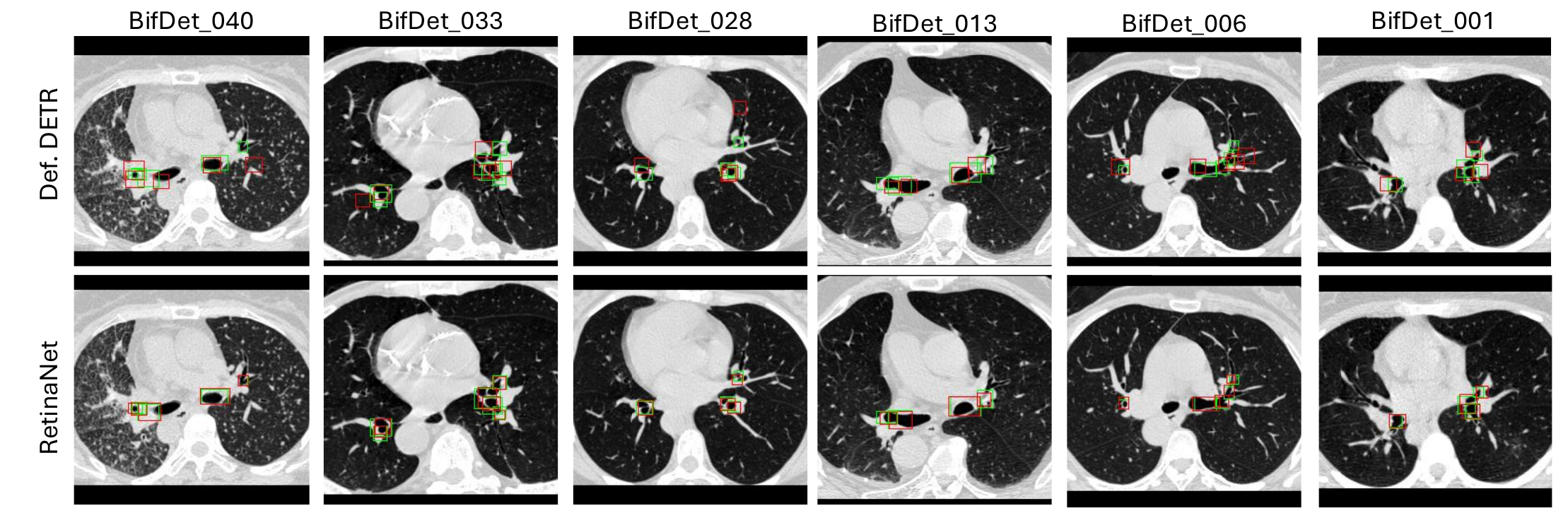}
\caption{Qualitative results comparing Deformable DETR (top row) and RetinaNet (bottom row) bifurcation detection performance (min. box size = 10). Green and red boxes respectively represent ground truths and predictions, and overlapping borders appear in yellow.
}
\label{fig:qualit}
\end{figure*}

\section{Results and Discussion}
\label{sec:results_discussion}

We report in Table \ref{tbl:quant_res} results of mAP and mAR along with $AP_{10}$, and $AR_{10}$ for a single IoU threshold of 0.1. This choice respects commonly used thresholds in medical imaging~\citep{retinunet_2020,baumgartner2021nndetection}. We report results for various minimal box dimension values, to focus on different bifurcation generations and hence sizes.

Overall, RetinaNet consistently outperforms DETR across all metrics and bifurcation sizes. As minimal bifurcation size increases, RetinaNet achieves substantial performance improvements, with mAP rising from \textbf{5.77\%} at 4 voxels to \textbf{67.92\%} at 10 voxels, and mAR increasing from \textbf{15.94\%} to \textbf{76.93\%}. DETR, in contrast, struggles across all thresholds, reaching only \textbf{2.35\% mAP} and \textbf{15.14\% mAR} at 10 voxels, reflecting its known limitations in detecting small, localized structures~\citep{melba:2023:003:wittmann}.

At low IoU threshold for the evaluated detections, RetinaNet also exhibits superior performance versus DETR as reflected in $\text{AP}_{10}$ and $\text{AR}_{10}$. At minimal bifurcation size of 4 voxels, its $\text{AP}_{10}$ is \textbf{10.60\%}, increasing to \textbf{74.66\%} at 10 voxels. Similarly, $\text{AR}_{10}$ shows a marked improvement from \textbf{23.90\%} to \textbf{83.67\%}. RetinaNet effectively identifies larger bifurcations even when localization precision requirements are relaxed, but its performance remains size-dependent. 
As expected, both models show decreasing performance with increasing IoU thresholds, with RetinaNet retaining relatively strong detection capabilities. For instance, at minimal bifurcation size of 10 voxels, RetinaNet achieves $\text{AP}_{25} = \mathbf{73.08\%}$ and $\text{AP}_{50} = \mathbf{47.96\%}$, while DETR drops sharply to $\text{AP}_{25} = \mathbf{1.37\%}$ and $\text{AP}_{50} = \mathbf{0.01\%}$. This confirms DETR’s difficulty in precise localization of bifurcations, especially at stricter overlap requirements.

Qualitative results (Figure~\ref{fig:qualit}) bring more insights on major differences between the two tested methods. RetinaNet returns less false positive boxes and seems to better approximate the real shape of the ground-truth box (e.g. square versus elongated). 

In terms of future work, our current annotations consist of bounding boxes marking only the spatial locations of bifurcations, without any associated semantic labels. This can be naturally extended to include anatomical information, such as branch-generation labels (e.g., trachea, main bronchi, lobar bronchi, and distal branches), enabling finer-grained analysis and hierarchical modeling of the airway tree.
Extraction of geometrical and morphological features at the annotated bifurcations remains to be done, such as branching angles, diameters, and shape asymmetries, which have been shown to be clinically relevant in prior works~\citep{quant_Cheung2882, appmed_fibrosis2021}. A detailed overview of these parameters and their definitions is provided in Appendix~\ref{appendix:bif_char}.

%\textbf{Technical Research Impact.} 
In terms of potential technical impact, BifDet can facilitate a wide range of current exciting research directions including anatomical tree matching~\citep{anattreematch_tschirren2005,anattreematch_graham2006}, graph-based methods for airway network extraction~\citep{graph_SELVAN2020,graph_Bauer2015} and airway tree anatomical labeling~\citep{graph_yu2023,graph_xie2022}. 
BifDet's potential extends to anomaly detection in bifurcation structures affected by specific diseases~\citep{appmed_fibrosis2021}
, shape analysis for understanding morphological variations~\citep{appmed_fibrosis2021}, and topological airway tree analysis~\citep{topo_Grothausmann2015}. Moreover, BifDet can be a valuable asset in boosting the training of automated airway segmentation models by guiding toward topologically-correct segmentation masks and serve in an additional performance metric of airway segmentation results. In the realm of computational fluid dynamics (CFD), BifDet can contribute to design more accurate ventilation simulations and respiratory drug delivery models by providing accurate locations of the airway bifurcations ~\citep{cfd_Bertolini2024, cfd_HOFMANN2011, cfd_kim2021, cfd_KOLANJIYIL2017, cfd_LONGEST2012}.

In terms of potential clinical impact, the quantification of airway bifurcation parameters enabled by BifDet has the potential to advance our biologic and clinical understanding of airway tree structure in lung diseases. This includes  bifurcation wall thickness, lumen area, branching patterns (angles, ratio, lengths, diameter, curvature, symmetry), and better assessing in between airway segments tapering (both inter-branch and intra-branch).
These properties can reflect developmental processes, influence disease susceptibility, and serve as biomarkers of respiratory disease risk and progression~\citep{quant_Cheung2882, appmed_copd2018, appmed_bronchiectasis2017, appmed_emphysema2014, medair_bokov2014}, potentially leading to earlier detection and more effective treatment strategies. By providing detailed localised information on parent and daughter branches within the airway tree, BifDet could enable a comprehensive understanding of individual anatomies, with potential for disease etiotyping and risk stratification. The ability to relate bifurcation morphology to specific pathologies has the potential to advance personalized approaches in respiratory care, paving the way for tailored interventions and improved patient outcomes.

\section{Conclusion}
\label{sec:conclusion}
In this paper, we introduced BifDet, the first publicly-available annotated dataset dedicated to 3D airway bifurcation detection.  
We formulated the annotation problem as a 3D bounding box delineation task and tested two deep-learning detection architectures using using our ground-truth annotations. 

Our detailed statistics on the annotated bounding boxes illustrate challenges raised by the intrinsic complex and variable nature of the airway tree, the high number of small bifurcations in close proximity to each others, and the inherent added complexity when relying on manually-annotated CT scans, which vary greatly in spatial coverage of the smaller airways either due to image quality of manual segmentation extent. 
 
Our quantitative and qualitative bifurcation detection results provide a concrete use-case  for our annotated dataset, with good baseline benchmark performance on larger airways. They also confirm remaining challenges for the densely-packed smaller airways, for which our annotations can be used for further research.
Indeed automated airway branch detection on lung CT scans could serve as guidelines to full airway segmentation. It could also be used as a stand-alone tool sufficient to extract airway biomarkers used by clinicians for diagnosis or treatment response evaluation.

We are planning the next version of BifDet, which will include disease cases from patients with conditions such as fibrosis and COPD, expanding beyond the current focus on healthy airways. This expanded dataset will also incorporate data from other cohorts, enhancing its diversity and generalizability, and opening doors for broader applications in respiratory health research.

\bibliography{refs}

@ARTICLE{exact09,
  author={Lo, Pechin and others},
  journal={IEEE Transactions on Medical Imaging}, 
  title={Extraction of Airways From CT (EXACT'09)}, 
  year={2012},
  volume={31},
  number={11},
  pages={2093-2107},}

@article{atm22,
title = {Multi-site, Multi-domain Airway Tree Modeling},
journal = {Medical Image Analysis},
volume = {90},
pages = {102957},
year = {2023},
issn = {1361-8415},
author = {Minghui Zhang and others},
}

@InProceedings{bas_dataset,
author="Qin, Yulei
and others",
editor="Martel, Anne L.
and Abolmaesumi, Purang
and Stoyanov, Danail
and Mateus, Diana
and Zuluaga, Maria A.
and Zhou, S. Kevin
and Racoceanu, Daniel
and Joskowicz, Leo",
title="Learning Bronchiole-Sensitive Airway Segmentation CNNs by Feature Recalibration and Attention Distillation",
booktitle="Medical Image Computing and Computer Assisted Intervention -- MICCAI 2020",
year="2020",
publisher="Springer International Publishing",
address="Cham",
pages="221--231",
isbn="978-3-030-59710-8"
}

@article{appmed_asthma2019,
   author = {Brody H. Foy and others},
   issn = {15354970},
   issue = {8},
   journal = {American Journal of Respiratory and Critical Care Medicine},
   month = {10},
   pages = {982-991},
   pmid = {31106566},
   publisher = {American Thoracic Society},
   title = {Lung computational models and the role of the small airways in asthma},
   volume = {200},
   year = {2019},
}

@article{appmed_bronchiectasis2017,
   author = {Wieying Kuo and others},
   issn = {14321084},
   issue = {11},
   journal = {European Radiology},
   month = {11},
   pages = {4680-4689},
   pmid = {28523349},
   publisher = {Springer Verlag},
   title = {Diagnosis of bronchiectasis and airway wall thickening in children with cystic fibrosis: Objective airway-artery quantification},
   volume = {27},
   year = {2017},
}

@article{appmed_copd2018,
   author = {Joshua Gawlitza and others},
   issn = {18727727},
   journal = {European Journal of Radiology},
   month = {7},
   pages = {87-93},
   pmid = {29857872},
   publisher = {Elsevier Ireland Ltd},
   title = {Finding the right spot: Where to measure airway parameters in patients with COPD},
   volume = {104},
   year = {2018},
}

@article{appmed_copd2018_2,
   author = {Benjamin M. Smith and others},
   issn = {10916490},
   issue = {5},
   journal = {Proceedings of the National Academy of Sciences of the United States of America},
   month = {1},
   pages = {E974-E981},
   pmid = {29339516},
   publisher = {National Academy of Sciences},
   title = {Human airway branch variation and chronic obstructive pulmonary disease},
   volume = {115},
   year = {2018},
}

@article{appmed_emphysema2014,
   author = {Misuzu Yahaba and others},
   issn = {18727727},
   issue = {6},
   journal = {European Journal of Radiology},
   month = {6},
   pages = {1022-1028},
   pmid = {24703520},
   publisher = {Elsevier Ireland Ltd},
   title = {The effects of emphysema on airway disease: Correlations between multi-detector CT and pulmonary function tests in smokers},
   volume = {83},
   year = {2014},
}

@article{appmed_fibrosis2021,
  title={Development of characteristic airway bifurcations in cystic fibrosis},
  author={Bass, Karl and others},
  journal={Aerosol Science and Technology},
  volume={55},
  number={10},
  pages={1143--1164},
  year={2021},
  publisher={Taylor \& Francis}
}

@ARTICLE{airseg_zhang2024,
  author={Zhang, Minghui and Gu, Yun},
  journal={IEEE Journal of Biomedical and Health Informatics}, 
  title={Towards Connectivity-Aware Pulmonary Airway Segmentation}, 
  year={2024},
  volume={28},
  number={1},
  pages={321-332}}

@ARTICLE{airseg_zheng2021,
  author={Zheng, Hao and others},
  journal={IEEE Transactions on Medical Imaging}, 
  title={Alleviating Class-Wise Gradient Imbalance for Pulmonary Airway Segmentation}, 
  year={2021},
  volume={40},
  number={9},
  pages={2452-2462}}

@InProceedings{airseg_qin2019,
author="Qin, Yulei
and others",
title="AirwayNet: A Voxel-Connectivity Aware Approach for Accurate Airway Segmentation Using Convolutional Neural Networks",
booktitle="Medical Image Computing and Computer Assisted Intervention -- MICCAI 2019",
year="2019",
publisher="Springer International Publishing",
address="Cham",
pages="212--220",
isbn="978-3-030-32226-7"
}

@article{medair_bokov2014,
title = {Homothety ratio of airway diameters and site of airway resistance in healthy and COPD subjects},
journal = {Respiratory Physiology \& Neurobiology},
volume = {191},
pages = {38-43},
year = {2014},
issn = {1569-9048},
author = {Plamen Bokov and others}
}

@article{cfd_bass2021,
author = {Karl Bass and others},
title = {Development of characteristic airway bifurcations in cystic fibrosis},
journal = {Aerosol Science and Technology},
volume = {55},
number = {10},
pages = {1143--1164},
year = {2021},
publisher = {Taylor \& Francis},
}

@InProceedings{cfd_Bertolini2024,
author="Bertolini, Michele
and others",
title="Towards Parametric Modelling of Human Bronchial Tree for Computational Fluid Dynamics",
booktitle="Design Tools and Methods in Industrial Engineering III",
year="2024",
publisher="Springer Nature Switzerland",
address="Cham",
pages="196--203",
isbn="978-3-031-58094-9"
}

@Article{cfd_kim2021,
AUTHOR = {Kim, Jongwon and Pidaparti, Ramana M.},
TITLE = {Computational Analysis of Lung and Isolated Airway Bifurcations under Mechanical Ventilation and Normal Breathing},
JOURNAL = {Fluids},
VOLUME = {6},
YEAR = {2021},
NUMBER = {11},
ARTICLE-NUMBER = {388},
ISSN = {2311-5521},
}

@article{cfd_KOLANJIYIL2017,
title = {Computational analysis of aerosol-dynamics in a human whole-lung airway model},
journal = {Journal of Aerosol Science},
volume = {114},
pages = {301-316},
year = {2017},
issn = {0021-8502},
doi = {https://doi.org/10.1016/j.jaerosci.2017.10.001},
author = {Arun V. Kolanjiyil and Clement Kleinstreuer}
}

@article{cfd_HOFMANN2011,
title = {Modelling inhaled particle deposition in the human lung—A review},
journal = {Journal of Aerosol Science},
volume = {42},
number = {10},
pages = {693-724},
year = {2011},
issn = {0021-8502},
author = {Werner Hofmann},
}

@article{cfd_LONGEST2012,
title = {In silico models of aerosol delivery to the respiratory tract — Development and applications},
journal = {Advanced Drug Delivery Reviews},
volume = {64},
number = {4},
pages = {296-311},
year = {2012},
note = {Computational and Visualization approaches in Respiratory Delivery},
issn = {0169-409X},
author = {P. Worth Longest and Landon T. Holbrook},
}

@article{anattreematch_tschirren2005,
   author = {Juerg Tschirren and others},
   issn = {02780062},
   issue = {12},
   journal = {IEEE Transactions on Medical Imaging},
   month = {12},
   pages = {1540-1547},
   pmid = {16353371},
   title = {Matching and anatomical labeling of human airway tree},
   volume = {24},
   year = {2005},
}

@article{anattreematch_graham2006,
   author = {Michael W. Graham and William E. Higgins},
   isbn = {0780395778},
   journal = {2006 3rd IEEE International Symposium on Biomedical Imaging: From Nano to Macro - Proceedings},
   pages = {109-112},
   title = {Optimal graph-theoretic approach to 3D anatomical tree matching},
   year = {2006},
}

@article{graph_SELVAN2020,
title = {Graph refinement based airway extraction using mean-field networks and graph neural networks},
journal = {Medical Image Analysis},
volume = {64},
pages = {101751},
year = {2020},
issn = {1361-8415},
author = {Raghavendra Selvan and others},
}

@ARTICLE{graph_Bauer2015,
  author={Bauer, Christian and others},
  journal={IEEE Transactions on Medical Imaging}, 
  title={Graph-Based Airway Tree Reconstruction From Chest CT Scans: Evaluation of Different Features on Five Cohorts}, 
  year={2015},
  volume={34},
  number={5},
  pages={1063-1076}}

@ARTICLE{graph_yu2023,
  author={Yu, Weihao and others},
  journal={IEEE Transactions on Medical Imaging}, 
  title={TNN: Tree Neural Network for Airway Anatomical Labeling}, 
  year={2023},
  volume={42},
  number={1},
  pages={103-118},}

@misc{graph_xie2022,
      title={Structure and position-aware graph neural network for airway labeling}, 
      author={Weiyi Xie and others},
      year={2022},
      eprint={2201.04532},
      archivePrefix={arXiv},
      primaryClass={cs.CV}
}

@article{topo_Grothausmann2015,
author = {Grothausmann, Roman and others},
year = {2015},
month = {02},
pages = {127010},
title = {Method for 3D Airway Topology Extraction},
volume = {2015},
journal = {Computational and mathematical methods in medicine},
}

@article {quant_Cheung2882,
	author = {Cheung, W K and others},
	title = {Airway inter-tapering and tortuosity predict mortality in idiopathic pulmonary fibrosis},
	volume = {60},
	number = {suppl 66},
	elocation-id = {2882},
	year = {2022},
	publisher = {European Respiratory Society},
	issn = {0903-1936},
	eprint = {https://erj.ersjournals.com/content},
	journal = {European Respiratory Journal}
}

@inproceedings{vessbif_zhao2014,
   author = {Mengliu Zhao and Ghassan Hamarneh},
   journal = {2014 IEEE 11th International Symposium on Biomedical Imaging (ISBI)},
   pages = {421-424},
   title = {Bifurcation detection in 3D vascular images using novel features and random forest},
   year = {2014},
}

@article{vessbif_shang2011,
   author = {Yanfeng Shang and others},
   issn = {00189294},
   issue = {4},
   journal = {IEEE Transactions on Biomedical Engineering},
   month = {4},
   pages = {1023-1032},
   pmid = {21138795},
   title = {Vascular active contour for vessel tree segmentation},
   volume = {58},
   year = {2011},
}

@article{vessbif_rafic2023,
title = {Using deep learning for an automatic detection and classification of the vascular bifurcations along the Circle of Willis},
journal = {Medical Image Analysis},
volume = {89},
pages = {102919},
year = {2023},
issn = {1361-8415},
author = {Rafic Nader and Romain Bourcier and Florent Autrusseau}
}

@article{vessbif_nouri2020,
title = {Characterization of 3D bifurcations in micro-scan and MRA-TOF images of cerebral vasculature for prediction of intra-cranial aneurysms},
journal = {Computerized Medical Imaging and Graphics},
volume = {84},
pages = {101751},
year = {2020},
issn = {0895-6111},
author = {A. Nouri and others}
}

@inproceedings{vessbif_zhao2017,
   author = {Mengliu Zhao and Ghassan Hamarneh},
   isbn = {9781538628188},
   journal = {Proceedings - 2017 14th Conference on Computer and Robot Vision, CRV 2017},
   month = {7},
   pages = {124-130},
   publisher = {Institute of Electrical and Electronics Engineers Inc.},
   title = {Bifurcation Localization in 3D Images via Evolutionary Geometric Deformable Templates},
   volume = {2018-January},
   year = {2017},
}

@INPROCEEDINGS{vessbif_wu2016,
  author={Wu, Aaron and others},
  booktitle={2016 IEEE 13th International Symposium on Biomedical Imaging (ISBI)}, 
  title={Deep vessel tracking: A generalized probabilistic approach via deep learning}, 
  year={2016},
  volume={},
  number={},
  pages={1363-1367}}

@InProceedings{airbif_zhao2018,
author="Zhao, Mengliu
and Hamarneh, Ghassan",
title="TreeNet: Multi-loss Deep Learning Network to Predict Branch Direction for Extracting 3D Anatomical Trees",
booktitle="Deep Learning in Medical Image Analysis and Multimodal Learning for Clinical Decision Support",
year="2018",
publisher="Springer International Publishing",
address="Cham",
pages="47--55",
isbn="978-3-030-00889-5"
}

@inproceedings{airbif_zhao2019,
   author = {Mengliu Zhao and Ghassan Hamarneh},
   isbn = {9783030326913},
   issn = {16113349},
   journal = {Lecture Notes in Computer Science (including subseries Lecture Notes in Artificial Intelligence and Lecture Notes in Bioinformatics)},
   pages = {637-645},
   publisher = {Springer},
   title = {Tree-LSTM: Using LSTM to Encode Memory in Anatomical Tree Prediction from 3D Images},
   volume = {11861 LNCS},
   year = {2019},
}

@article{airbif_wang2020,
   author = {Manyang Wang and others},
   issn = {17410444},
   issue = {9},
   journal = {Medical and Biological Engineering and Computing},
   month = {9},
   pages = {2009-2024},
   publisher = {Springer},
   title = {Automated labeling of the airway tree in terms of lobes based on deep learning of bifurcation point detection},
   volume = {58},
   year = {2020},
}

@InProceedings{retinunet_2020,
  title = 	 {{Retina U-Net: Embarrassingly Simple Exploitation of Segmentation Supervision for Medical Object Detection}},
  author =       {Jaeger, Paul F. and others},
  booktitle = 	 {Proceedings of the Machine Learning for Health NeurIPS Workshop},
  pages = 	 {171--183},
  year = 	 {2020},
  volume = 	 {116},
  series = 	 {Proceedings of Machine Learning Research},
  month = 	 {13 Dec},
  publisher =    {PMLR},
}

@InProceedings{medobjdetection_ma2021,
author="Ma, Xinghua
and others",
title="Transformer Network for Significant Stenosis Detection in CCTA of Coronary Arteries",
booktitle="Medical Image Computing and Computer Assisted Intervention -- MICCAI 2021",
year="2021",
publisher="Springer International Publishing",
address="Cham",
pages="516--525",
isbn="978-3-030-87231-1"
}

@InProceedings{retinanet_Lin2017,
author = {Lin, Tsung-Yi and others},
title = {Focal Loss for Dense Object Detection},
booktitle = {Proceedings of the IEEE International Conference on Computer Vision (ICCV)},
month = {Oct},
year = {2017}
}

@inproceedings{frcnn_ren2015,
 author = {Ren, Shaoqing and others},
 booktitle = {Advances in Neural Information Processing Systems},
 pages = {},
 publisher = {Curran Associates, Inc.},
 title = {Faster R-CNN: Towards Real-Time Object Detection with Region Proposal Networks},
 volume = {28},
 year = {2015}
}

@InProceedings{detr_carion2020,
author="Carion, Nicolas
and others",
title="End-to-End Object Detection with Transformers",
booktitle="Computer Vision -- ECCV 2020",
year="2020",
publisher="Springer International Publishing",
address="Cham",
pages="213--229",
isbn="978-3-030-58452-8"
}

@inproceedings{defdetr_zhu2021,
  author       = {Xizhou Zhu and
                  others},
  title        = {Deformable {DETR:} Deformable Transformers for End-to-End Object Detection},
  booktitle    = {9th International Conference on Learning Representations, {ICLR} 2021,
                  Virtual Event, Austria, May 3-7, 2021},
  publisher    = {OpenReview.net},
  year         = {2021}
}

@inproceedings{medobjdetection_Zlocha2019,
  author    = {Martin Zlocha and Qi Dou and Ben Glocker},
  title     = {Improving RetinaNet for CT Lesion Detection with Dense Masks from Weak RECIST Labels},
  booktitle = {Medical Image Computing and Computer Assisted Intervention - MICCAI 2019 - 22nd International Conference, Shenzhen, China, October 13-17, 2019, Proceedings, Part VI},
  series    = {Lecture Notes in Computer Science},
  volume    = {11769},
  pages     = {402--410},
  year      = {2019},
  publisher = {Springer}
}

@inproceedings{medobjdetection_shen2021,
   author = {Zhiqiang Shen and others},
   isbn = {9781665409506},
   journal = {2021 7th International Conference on Computer and Communications, ICCC 2021},
   pages = {1757-1761},
   publisher = {Institute of Electrical and Electronics Engineers Inc.},
   title = {COTR: Convolution in Transformer Network for End to End Polyp Detection},
   year = {2021},
}

@inproceedings{medobjdetection_yan2018,
  title={3D Context Enhanced Region-based Convolutional Neural Network for End-to-End Lesion Detection},
  author={Ke Yan and Mohammadhadi Bagheri and Ronald M. Summers},
  booktitle={International Conference on Medical Image Computing and Computer-Assisted Intervention},
  year={2018}
}

@article{medobjdetection_wu2020,
title = {Automatic detection of coronary artery stenosis by convolutional neural network with temporal constraint},
journal = {Computers in Biology and Medicine},
volume = {118},
pages = {103657},
year = {2020},
issn = {0010-4825},
author = {Wei Wu and others}
}

@article{medobjdetection_xie2019,
title = {Automated pulmonary nodule detection in CT images using deep convolutional neural networks},
journal = {Pattern Recognition},
volume = {85},
pages = {109-119},
year = {2019},
issn = {0031-3203},
author = {Hongtao Xie and others},
}

@INPROCEEDINGS{medobjdetection_zhu2018,
  author={Zhu, Wentao and others},
  booktitle={2018 IEEE Winter Conference on Applications of Computer Vision (WACV)}, 
  title={DeepLung: Deep 3D Dual Path Nets for Automated Pulmonary Nodule Detection and Classification}, 
  year={2018},
  volume={},
  number={},
  pages={673-681}}

@article{melba:2023:003:wittmann,
    title = "Focused Decoding Enables 3D Anatomical Detection by Transformers",
    author = "Wittmann, Bastian and others",
    journal = "Machine Learning for Biomedical Imaging",
    volume = "2",
    issue = "February 2023 issue",
    year = "2023",
    pages = "72--95",
    issn = "2766-905X",
}

@inproceedings{lin2017feature,
  title={Feature pyramid networks for object detection},
  author={Lin, Tsung-Yi and others},
  booktitle={Proceedings of the IEEE conference on computer vision and pattern recognition},
  pages={2117--2125},
  year={2017}
}

@inproceedings{girshick2015fast,
  title={Fast r-cnn},
  author={Girshick, Ross},
  booktitle={Proceedings of the IEEE international conference on computer vision},
  pages={1440--1448},
  year={2015}
}

@article{vaswani2017attention,
  title={Attention is all you need},
  author={Vaswani, Ashish and others},
  journal={Advances in neural information processing systems},
  volume={30},
  year={2017}
}

@inproceedings{zhang2020bridging,
  title={Bridging the gap between anchor-based and anchor-free detection via adaptive training sample selection},
  author={Zhang, Shifeng and others},
  booktitle={Proceedings of the IEEE/CVF conference on computer vision and pattern recognition},
  pages={9759--9768},
  year={2020}
}

@inproceedings{loshchilov2018decoupled,
  title={Decoupled Weight Decay Regularization},
  author={Loshchilov, Ilya and Hutter, Frank},
  booktitle={International Conference on Learning Representations},
  year={2018}
}

@ARTICLE{lungsmaskhofmanninger2020,
  title     = "Automatic lung segmentation in routine imaging is primarily a
               data diversity problem, not a methodology problem",
  author    = "Hofmanninger, J. and others",
  journal   = "Eur. Radiol. Exp.",
  publisher = "Springer Science and Business Media LLC",
  volume    =  4,
  number    =  1,
  pages     = "50",
  month     =  aug,
  year      =  2020,
  copyright = "https://creativecommons.org/licenses/by/4.0",
  language  = "en"
}

@inproceedings{lin2014microsoft,
  title={Microsoft coco: Common objects in context},
  author={Lin, Tsung-Yi and others},
  booktitle={Computer Vision--ECCV 2014: 13th European Conference, Zurich, Switzerland, September 6-12, 2014, Proceedings, Part V 13},
  pages={740--755},
  year={2014},
  organization={Springer}
}

@inproceedings{baumgartner2021nndetection,
  title={nnDetection: a self-configuring method for medical object detection},
  author={Baumgartner, Michael and others},
  booktitle={Medical Image Computing and Computer Assisted Intervention--MICCAI 2021: 24th International Conference, Strasbourg, France, September 27--October 1, 2021, Proceedings, Part V 24},
  pages={530--539},
  year={2021},
  organization={Springer}
}

@article{bif_grothausmann2015,
   author = {Roman Grothausmann and others},
   issue = {1},
   journal = {Computational and Mathematical Methods in Medicine},
   pages = {127010},
   title = {Method for 3D Airway Topology Extraction},
   volume = {2015},
   year = {2015},
}

@article{Tuinenburg2011,
   author = {Joan C. Tuinenburg and others},
   issn = {15730743},
   issue = {2},
   journal = {International Journal of Cardiovascular Imaging},
   pages = {167-174},
   pmid = {21327913},
   publisher = {Kluwer Academic Publishers},
   title = {Dedicated bifurcation analysis: Basic principles},
   volume = {27},
   year = {2011},
}

@article{Chassagnon2016,
   author = {Guillaume Chassagnon and others},
   issn = {15271323},
   issue = {2},
   journal = {Radiographics},
   month = {3},
   pages = {358-373},
   pmid = {26824513},
   publisher = {Radiological Society of North America Inc.},
   title = {Tracheobronchial branching abnormalities: Lobe-based classification scheme},
   volume = {36},
   year = {2016},
}

@article{Niimi2003,
author = {Niimi, Akio and others},
title = {Relationship of Airway Wall Thickness to Airway Sensitivity and Airway Reactivity in Asthma},
journal = {American Journal of Respiratory and Critical Care Medicine},
volume = {168},
number = {8},
pages = {983-988},
year = {2003},
}

@article{Van_de_Moortele2019,
title={Airway morphology and inspiratory flow features in the early stages of Chronic Obstructive Pulmonary Disease},
author={Van de Moortele, T. and others},
journal={Clin Biomech (Bristol, Avon)},
volume={66},
pages={60-65},
year={2019},
month={Jun},
epub={2017-11-16},
pmid={29169684},
pmcid={PMC5955793}
}

@article{Bense2004,
author = {Bense, L. and others},
year = {2004},
month = {09},
pages = {355-384},
title = {SYMMETRY, STRUCTURE, HIERARCHY AND TOPOLOGY IN A BRONCHIAL TREE STRUCTURAL MODEL OF THE LUNG},
volume = {14-15},
journal = {Symmetry: Culture and Science}
}

@InProceedings{sg_net2024,
author="Tan, Zimeng
and Feng, Jianjiang
and Zhou, Jie",
title="SGNet: Structure-Aware Graph-Based Network for Airway Semantic Segmentation",
booktitle="Medical Image Computing and Computer Assisted Intervention -- MICCAI 2021",
year="2021",
publisher="Springer International Publishing",
address="Cham",
pages="153--163",
isbn="978-3-030-87193-2"
}

@article{aiib23_ds,
title = {Hunting imaging biomarkers in pulmonary fibrosis: Benchmarks of the AIIB23 challenge},
journal = {Medical Image Analysis},
volume = {97},
pages = {103253},
year = {2024},
issn = {1361-8415},
author = {Yang Nan and others},
}

@ARTICLE{bronchotrack,
  author={Tian, Qingyao and others},
  journal={IEEE Transactions on Medical Imaging}, 
  title={BronchoTrack: Airway Lumen Tracking for Branch-Level Bronchoscopic Localization}, 
  year={2025},
  volume={44},
  number={3},
  pages={1321-1333}}

@article{LEE1994462,
title = {Building Skeleton Models via 3-D Medial Surface Axis Thinning Algorithms},
journal = {CVGIP: Graphical Models and Image Processing},
volume = {56},
number = {6},
pages = {462-478},
year = {1994},
issn = {1049-9652},
author = {T.C. Lee and R.L. Kashyap and C.N. Chu},
}

@article{SAHA20163,
title = {A survey on skeletonization algorithms and their applications},
journal = {Pattern Recognition Letters},
volume = {76},
pages = {3-12},
year = {2016},
note = {Special Issue on Skeletonization and its Application},
issn = {0167-8655},
author = {Punam K. Saha and Gunilla Borgefors and Gabriella {Sanniti di Baja}},
}

@article{NEJMoa2414059,
author = {Robert J. Lentz  and others},
title = {Navigational Bronchoscopy or Transthoracic Needle Biopsy for Lung Nodules},
journal = {New England Journal of Medicine},
volume = {392},
number = {21},
pages = {2100-2112},
year = {2025},
doi = {10.1056/NEJMoa2414059},
URL = {https://www.nejm.org/doi/full/10.1056/NEJMoa2414059},
eprint = {https://www.nejm.org/doi/pdf/10.1056/NEJMoa2414059}
}

\appendix
\section{BifDet Access and Documentation}
\label{appendix:dataset_info}
\textbf{Documentation and Intended Uses.}
The BifDet dataset is specifically designed for the development and evaluation of 3D airway bifurcation detection algorithms. It can be used for various research tasks, such as anatomical tree matching, graph-based airway network extraction, airway tree anatomical labeling, anomaly detection, shape analysis, topological airway tree analysis, and ventilation simulations.

\textbf{URL to Download Page.}
During the review process, the BifDet dataset is available for download on the Google Drive platform (\href{https://drive.google.com/drive/folders/1vZOJlE9yX5QqzwyzsGW0j1-iQuaGKg13?usp=share_link}{LINK}). We plan to publish it on the Zenodo\footnote{\url{https://zenodo.org/}} platform with specific DOI.

\textbf{URL to the project GitHub.} 
% \faGithub\ 
\href{https://github.com/ali2066k/BifDet2024}{ BifDet2024}

\textbf{Author Statement and Data License.}
The authors of the BifDet dataset bear all responsibility for any violations of rights or ethical considerations related to the data. The dataset is released under a Creative Commons Attribution-NonCommercial-ShareAlike 4.0 International (CC BY-NC-SA 4.0) License. The original CT scan data used for annotation is sourced from the ATM22 dataset, and its terms of use can be found at the following link: (https://atm22.grand-challenge.org)

\textbf{Hosting, Licensing, and Maintenance Plan.}
The BifDet dataset will be hosted on a dedicated data sharing platform for ensuring its accessibility, long-term preservation, and facilitating the assignment of a Digital Object Identifier (DOI) for proper citation and reference upon publication. The authors are committed to maintaining the dataset by addressing any errors or inconsistencies that may be identified and by potentially expanding the dataset in the future to include pathological cases and data from diverse cohorts.

\section{Details on BifDet CT scans and Annotations}
\label{appendix:bifdet_ATM22}

\subsection*{A. CT Scan Characteristics and Case Metadata}
Table~\ref{tbl:bifdet_meta_info} provides details on the 42 BifDet cases, each corresponding to an anonymized patient scan from the ATM22 dataset. All scans have a consistent axial matrix size of $512 \times 512$ pixels and a slice thickness of 0.5 mm. The number of slices per scan ranges from 615 to 799, with an average of 723 in the full volumes. After lung field-of-view extraction, the average number of slices is reduced to 499.05, with a range between 307 and 633 slices. The axial voxel spacing is always isotropic, typically around 0.80 mm, with slight variations across cases ranging from 0.74 mm to 0.90 mm.
%The data type for all cases is int16, indicating that the pixel values are stored as 16-bit integers.

\begin{table*}[htbp]
    \centering
    % \scriptsize
    \caption{List of ATM22 scans annotated in BifDet. Last column indicates the split inin training subset (checkmark) and validation subset (hyphen).}
    \label{tbl:bifdet_meta_info}
    \resizebox{\linewidth}{!}{%
    \begin{tabular}{cccccccc}
        \toprule
        BifDet Case & ATM22 Name & Axial Dimension & Slice Count & Voxel Size (mm) & Slice Thickness (mm) & Train. & \# Boxes \\
        \midrule
BifDet\_001 & ATM\_001 & 512x512 & 679 & 0.82x0.82 & 0.50 & - & 140 \\ 
BifDet\_002 & ATM\_002 & 512x512 & 635 & 0.74x0.74 & 0.50 & \checkmark & 143 \\ 
BifDet\_003 & ATM\_003 & 512x512 & 799 & 0.78x0.78 & 0.50 & \checkmark & 164 \\ 
BifDet\_004 & ATM\_004 & 512x512 & 679 & 0.82x0.82 & 0.50 & \checkmark & 237 \\ 
BifDet\_005 & ATM\_005 & 512x512 & 741 & 0.90x0.90 & 0.50 & \checkmark & 135 \\ 
BifDet\_006 & ATM\_006 & 512x512 & 679 & 0.82x0.82 & 0.50 & - & 195 \\ 
BifDet\_007 & ATM\_007 & 512x512 & 799 & 0.78x0.78 & 0.50 & \checkmark & 177 \\ 
BifDet\_008 & ATM\_008 & 512x512 & 679 & 0.82x0.82 & 0.50 & \checkmark & 144 \\ 
BifDet\_009 & ATM\_009 & 512x512 & 799 & 0.78x0.78 & 0.50 & \checkmark & 149 \\ 
BifDet\_010 & ATM\_010 & 512x512 & 799 & 0.78x0.78 & 0.50 & \checkmark & 171 \\ 
BifDet\_011 & ATM\_011 & 512x512 & 799 & 0.78x0.78 & 0.50 & \checkmark & 231 \\ 
BifDet\_012 & ATM\_012 & 512x512 & 799 & 0.78x0.78 & 0.50 & \checkmark & 162 \\ 
BifDet\_013 & ATM\_013 & 512x512 & 755 & 0.84x0.84 & 0.50 & - & 218 \\ 
BifDet\_014 & ATM\_014 & 512x512 & 621 & 0.85x0.85 & 0.50 & \checkmark & 238 \\ 
BifDet\_015 & ATM\_015 & 512x512 & 799 & 0.78x0.78 & 0.50 & \checkmark & 166 \\ 
BifDet\_016 & ATM\_021 & 512x512 & 724 & 0.82x0.82 & 0.50 & \checkmark & 197 \\ 
BifDet\_017 & ATM\_022 & 512x512 & 655 & 0.78x0.78 & 0.50 & \checkmark & 255 \\ 
BifDet\_018 & ATM\_023 & 512x512 & 799 & 0.86x0.86 & 0.50 & - & 138 \\ 
BifDet\_019 & ATM\_024 & 512x512 & 657 & 0.78x0.78 & 0.50 & \checkmark & 242 \\ 
BifDet\_020 & ATM\_025 & 512x512 & 799 & 0.78x0.78 & 0.50 & \checkmark & 251 \\ 
BifDet\_021 & ATM\_026 & 512x512 & 679 & 0.82x0.82 & 0.50 & \checkmark & 117 \\ 
BifDet\_022 & ATM\_027 & 512x512 & 615 & 0.90x0.90 & 0.50 & \checkmark & \textbf{72} \\ 
BifDet\_023 & ATM\_028 & 512x512 & 703 & 0.78x0.78 & 0.50 & \checkmark & 142 \\ 
BifDet\_024 & ATM\_029 & 512x512 & 687 & 0.78x0.78 & 0.50 & \checkmark & 184 \\ 
BifDet\_025 & ATM\_030 & 512x512 & 765 & 0.82x0.82 & 0.50 & \checkmark & 218 \\ 
BifDet\_026 & ATM\_031 & 512x512 & 799 & 0.78x0.78 & 0.50 & \checkmark & 207 \\ 
BifDet\_027 & ATM\_032 & 512x512 & 705 & 0.82x0.82 & 0.50 & \checkmark & 177 \\ 
BifDet\_028 & ATM\_033 & 512x512 & 655 & 0.78x0.78 & 0.50 & - & 129 \\ 
BifDet\_029 & ATM\_034 & 512x512 & 769 & 0.78x0.78 & 0.50 & \checkmark & 150 \\ 
BifDet\_030 & ATM\_036 & 512x512 & 759 & 0.84x0.84 & 0.50 & \checkmark & 226 \\ 
BifDet\_031 & ATM\_037 & 512x512 & 799 & 0.78x0.78 & 0.50 & - & 132 \\ 
BifDet\_032 & ATM\_038 & 512x512 & 679 & 0.82x0.82 & 0.50 & \checkmark & 206 \\ 
BifDet\_033 & ATM\_039 & 512x512 & 799 & 0.78x0.78 & 0.50 & - & 270 \\ 
BifDet\_034 & ATM\_040 & 512x512 & 640 & 0.82x0.82 & 0.50 & \checkmark & 191 \\ 
BifDet\_035 & ATM\_041 & 512x512 & 799 & 0.78x0.78 & 0.50 & - & 162 \\ 
BifDet\_036 & ATM\_042 & 512x512 & 687 & 0.78x0.78 & 0.50 & \checkmark  & \textbf{274} \\ 
BifDet\_037 & ATM\_043 & 512x512 & 667 & 0.87x0.87 & 0.50 & \checkmark & 163 \\ 
BifDet\_038 & ATM\_044 & 512x512 & 679 & 0.82x0.82 & 0.50 & \checkmark &  123 \\ 
BifDet\_039 & ATM\_045 & 512x512 & 737 & 0.73x0.73 & 0.50 & \checkmark & 168 \\ 
BifDet\_040 & ATM\_046 & 512x512 & 679 & 0.82x0.82 & 0.50 & - & 131 \\ 
BifDet\_041 & ATM\_047 & 512x512 & 679 & 0.82x0.82 & 0.50 & \checkmark & 194 \\ 
BifDet\_042 & ATM\_048 & 512x512 & 697 & 0.78x0.78 & 0.50 & \checkmark & 133 \\ 
        \midrule
        & & 512x512 & 723.14 & 0.80x0.80 & 0.50 & & 179 \\
        \bottomrule
    \end{tabular}}
\end{table*}

\begin{figure*}[tbp]
\centering
\includegraphics[width=0.75\textwidth]{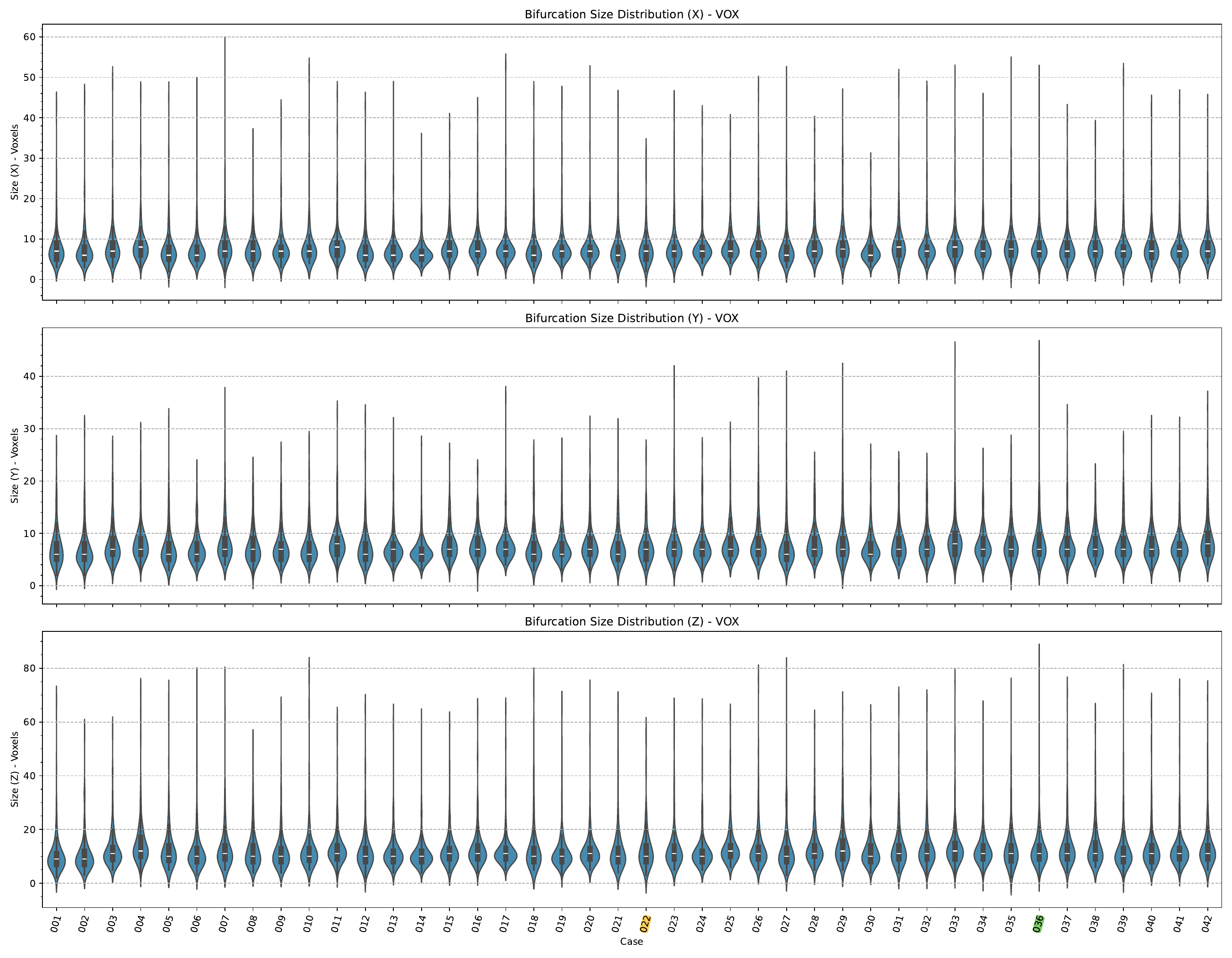}
\caption{Per-case violin plots of bifurcation bounding box sizes across the X, Y, and Z axes, shown in (a) physical units (mm) and (b) voxel units. Each violin represents a single CT scan case; wider regions indicate higher density of bifurcation sizes. The distributions reflect both inter-case anatomical variability and resolution heterogeneity, with sizes skewed toward smaller values.}
\label{fig:violin_vox}
\end{figure*}

\begin{figure*}[tbp]
\centering
\includegraphics[width=0.99\textwidth]{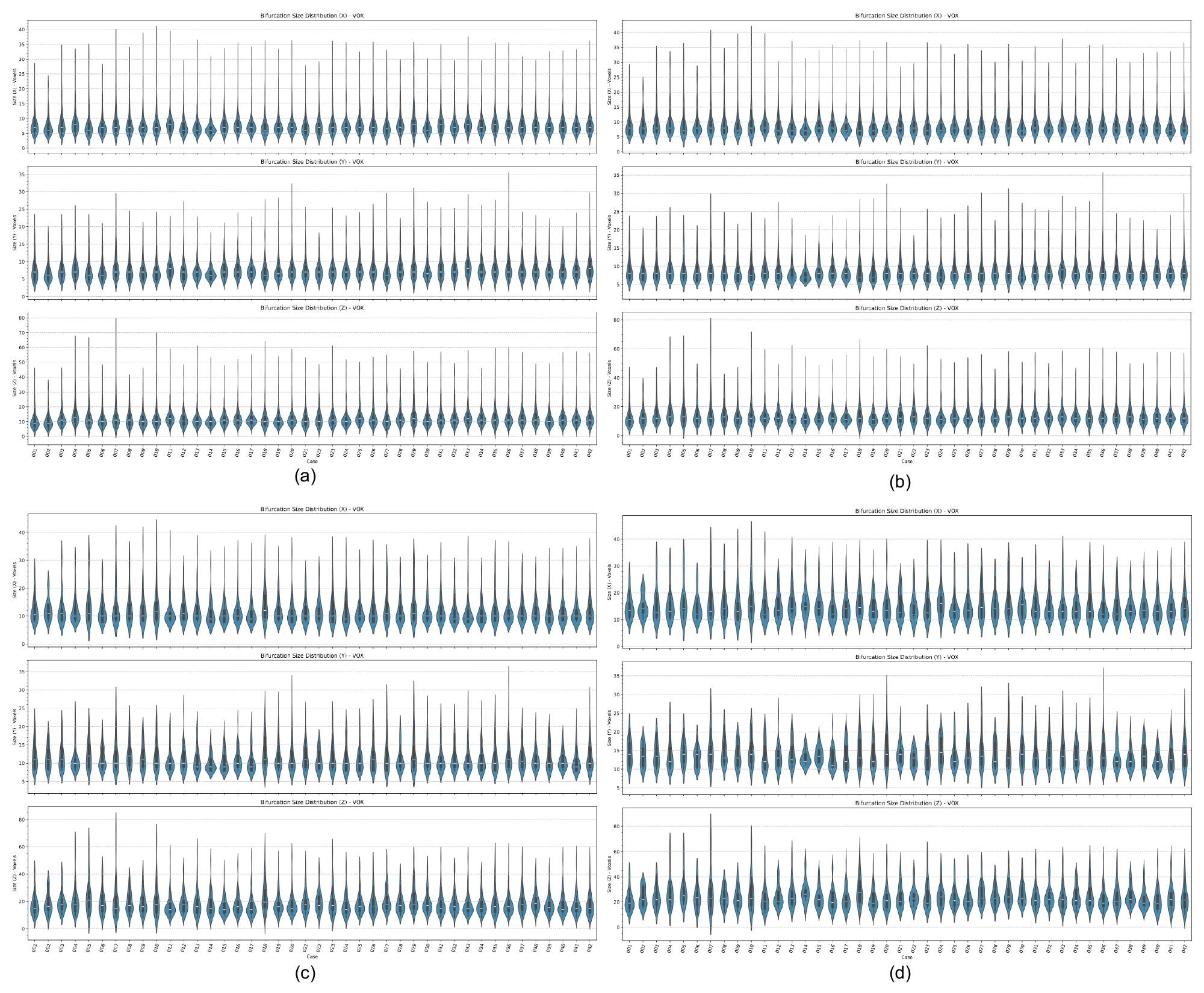}
\caption{Per-case distributions of bifurcation sizes along each spatial axis under varying minimum voxel size thresholds used during training: (a) 4 voxels, (b) 6 voxels, (c) 8 voxels, and (d) 10 voxels. The plots highlight anatomical variability and the prevalence of small bifurcations across cases.}
\label{fig:violin_formin}
\end{figure*}

To further characterize the anatomical variability, Figure~\ref{fig:violin_vox} presents per-case distributions of bifurcation sizes along each spatial axis, both in millimeters and voxels. The violin plots highlight substantial variation in size distributions across scans, with a majority of bifurcations skewed toward smaller sizes. This heterogeneity, combined with the dense clustering of small objects, reinforces the difficulty of 3D bifurcation detection in real clinical data. Figure~\ref{fig:violin_formin} illustrates the size distributions corresponding to the minimum box size used in each training scenario.

\subsection*{B. Distribution of Bifurcation Sizes and Resolution Effects}

\begin{figure*}[tbp]
\centering
\includegraphics[width=0.99\textwidth]{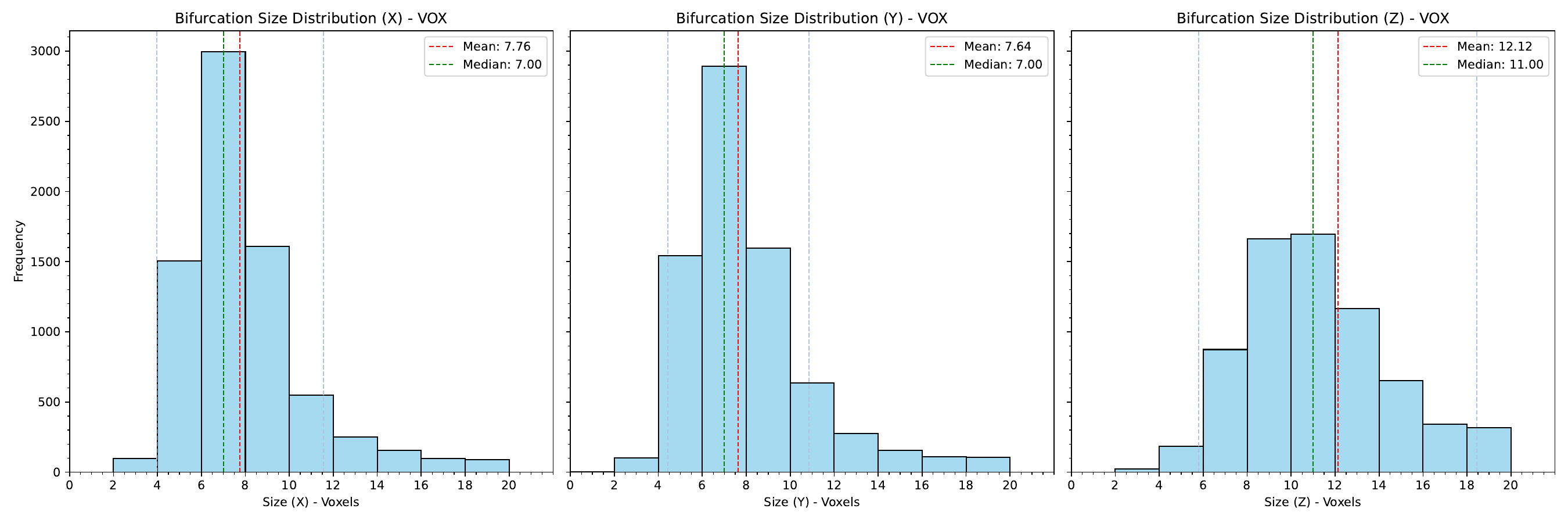}
\caption{
Histogram of bifurcation sizes along the X, Y, and Z directions in voxel units. All plots use fixed bin intervals of 2 voxels to ensure comparability. The Z-direction shows larger sizes on average. The drop in small sizes below 6 voxels reflects CT scam resolution limitations and partial volume effects.
}
\label{fig:histogram_vox_xyz}
\end{figure*}

\begin{figure*}[tbp]
\centering
\includegraphics[width=0.99\textwidth]{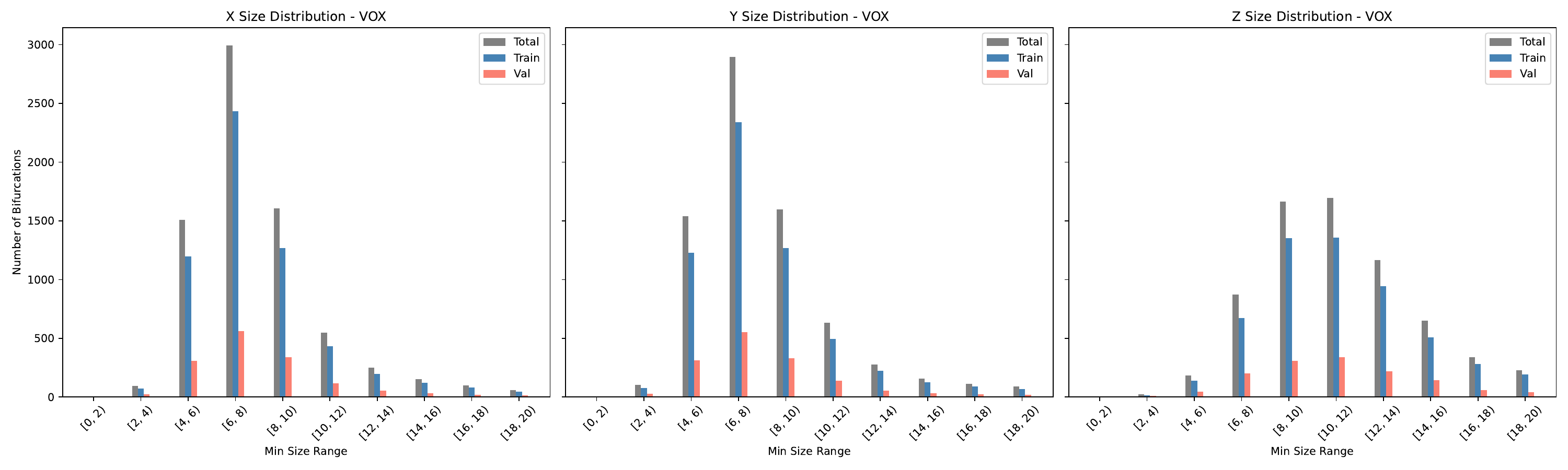}
\caption{
Grouped bar plots showing the number of bifurcations per size bin along the X, Y, and Z axes. Each bin contains three bars corresponding to the total dataset, training set, and validation set. The distributions are well aligned across splits, demonstrating consistent representation of bifurcation scales.
}
\label{fig:grouped_histograms}
\end{figure*}

To better understand the spatial characteristics of annotated bifurcations, we analyzed the size distribution along the X, Y, and Z axes across all cases. Figure~\ref{fig:histogram_vox_xyz} shows voxel-wise histograms using consistent bin edges ([0–2), [2–4), ..., [20–22)) for all three axes. These plots reveal a clear skew toward smaller bifurcations, with a drop in frequency for boxes smaller than 6 voxels. This reflects resolution limitations and voxel discretization effects, which hinder the consistent annotation of very small bifurcations.
The Z-direction distributions show systematically larger sizes compared to X and Y (0.5 mm slice thickness vs. ~0.8 mm in-plane). This has implications on the minimum detectable bifurcation size and influences the choice of training thresholds.
To assess the balance of size distributions across the data splits, we further computed grouped histograms for the training and validation sets (Figure~\ref{fig:grouped_histograms}). Each plot displays three adjacent bars per bin: total, training, and validation counts. The curves are well-aligned, confirming consistent representation across subsets.

\section{Motivation for Bifurcations annotated as Bounding Boxes}
\label{appendix:bif_char}

\begin{figure}[tbp]
\centering
\includegraphics[width=0.5\columnwidth]{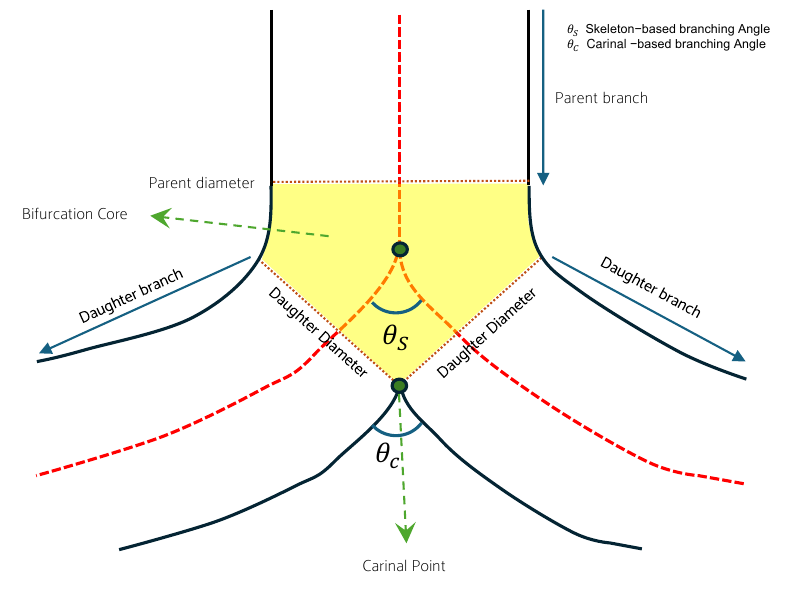}
\caption{A 2D Bifurcation Model depicting the parent branch, carinal point, and the two daughter branches along with branching angles $\theta_c$ and $\theta_S$. Inspired by~\citep{Tuinenburg2011}. 
}
\label{fig:bif_2d_model}
\end{figure}

Annotating an airway bifurcations as bounding boxes enables to encompass the parent branch, the division point (carinal point) and the two (or sometimes more) daughter branches resulting from the split. 
The number of  branching points with more than two daughters has been used as a biomarker for COPD in~\citep{appmed_copd2018_2}.
Parent and daughter branches can be described using distinct geometric and morphological parameters, such as lumen diameter, cross-sectional area, and branching angles. These characteristics are critical for quantifying bifurcation anatomy and understanding airflow dynamics. For example, Tuinenburg et al.\citep{Tuinenburg2011} proposed a 2D bifurcation model to define anatomical angles at carinal points (see Figure\ref{fig:bif_2d_model}). Furthermore, studies such as Bertolini et al.~\citep{appmed_fibrosis2021} have utilized branch-wise measurements—including diameters, lengths, and angles—to model airway remodeling in fibrotic lung disease, underscoring the clinical value of detailed bifurcation characterization.

The following geometric and morphological characteristics have been assessed in the analysis of airway bifurcations:

\begin{itemize}
    \item \textbf{Branching Angles:} Abnormal branching angles have been studied as indicative of underlying respiratory conditions such as bronchiectasis or COPD~\citep{Chassagnon2016}. The angles formed between the parent and daughter branches at the trachea carinal point are crucial for understanding airflow patterns and pressure distribution within the airways ~\citep{}. .
    \item \textbf{Diameter and Area:} The diameter and cross-sectional area of each branch are essential for assessing airflow capacity and resistance. Changes in these parameters can affect ventilation and gas exchange and may be observed in asthma or COPD~\citep{Van_de_Moortele2019}.
    \item \textbf{Wall Thickness:} The thickness of the airway wall can vary in different regions and may be altered in certain diseases, impacting airway compliance and responsiveness. Increased wall thickness is a hallmark of chronic inflammatory airway diseases~\citep{Niimi2003}.
    \item \textbf{Radius of Curvature:} The curvature of the airway branches, particularly at the carinal points, can affect airflow patterns and particle deposition. Excessive curvature may be seen in some congenital or acquired airway abnormalities.
    \item \textbf{Symmetry:} The degree of symmetry between the two daughter branches is relevant for understanding airflow distribution and potential asymmetries in ventilation~\citep{Bense2004}. 
    % Asymmetries may arise due to tumors or other obstructive lesions~\citep{}.
    \item \textbf{Carinal Shape:} The shape at the carinal point (e.g., sharp, blunt, Y-shaped) can influence airflow dynamics and the likelihood of particle impaction. Alterations in carinal shape can occur in various respiratory diseases.
\end{itemize}
 
By analyzing these bifurcation characteristics in both healthy and diseased individuals, researchers can gain a deeper understanding of how airway structure relates to respiratory function and disease pathogenesis, while avoiding to run cumbersome and error-prone full segmentations of the airway tree. In addition,  correctly detecting branching points from a binary segmentation mask of the airway tree is not trivial, as skeletonization algorithms are subject to artifacts and errors~\citep{bif_grothausmann2015}.

\section{Evaluation metrics}\label{appendix:eval_metrics}

To rigorously assess the performance of our 3D bifurcation detection models, we employ established metrics widely used in object detection tasks: Average Precision (AP) and Average Recall (AR). 
More specifically, we are using the versions of these metrics implemented for the COCO dataset and benchmark \citep{lin2014microsoft}.
These metrics offer a comprehensive evaluation of the model's ability to both accurately localize and classify bifurcations within 3D medical images.

\subsection{Intersection over Union (IoU)}
At the core of our evaluation methodology lies the Intersection over Union (IoU) criterion, which determines whether a predicted bounding box qualifies as a true positive detection. Given a predicted bounding box $\rmB_{(i,j)} = (\hat{\rvc}_{(i,j)}, \hat{w}_{(i,j)}, \hat{h}_{(i,j)}, \hat{d}_{(i,j)})$ and a ground truth bounding box $\rmG_{(i,j)} = (\rvc_{(i,j)}^*,w_{(i,j)}^*, h_{(i,j)}^*, d_{(i,j)}^*)$, the IoU is calculated as:

$$
\text{IoU}(\rmB{(i,j)}, \rmG{(i,j)}) = \frac{\text{Volume}(\rmB{(i,j)} \cap \rmG{(i,j)})}{\text{Volume}(\rmB{(i,j)} \cup \rmG{(i,j)})}
$$

where $\cap$ denotes the intersection and $\cup$ denotes the union of the two boxes.

\subsection{True Positives, False Positives, and False Negatives}
Based on the IoU criterion and a predefined threshold T, we classify predictions into three categories:

True Positive (TP): A predicted box $\rmB_{(i,j)}$ is considered a TP if there exists a ground truth box $\rmG_{(i,j)}$ such that $\text{IoU}(\rmB_{(i,j)}, \rmG_{(i,j)}) > T$.

False Positive (FP): A predicted box $\rmB_{(i,j)}$ is considered an FP if there does not exist any ground truth box $\rmG_{(i,j)}$ such that $\text{IoU}(\rmB_{(i,j)}, \rmG_{(i,j)}) > T$.

False Negative (FN): A ground truth box $\rmG_{(i,j)}$ is considered an FN if there does not exist any predicted box $\rmB_{(i,j)}$ such that $\text{IoU}(\rmB_{(i,j)}, \rmG_{(i,j)}) > T$.

\subsection{Precision and Recall}
For a given IoU threshold T, we calculate the precision and recall as:

$$
\text{Precision}(T) = \frac{\text{TP}(T)}{\text{TP}(T) + \text{FP}(T)}
$$

$$
\text{Recall}(T) = \frac{\text{TP}(T)}{\text{TP}(T) + \text{FN}(T)}
$$

\subsection{Mean Average Precision (mAP) and Recall (mAR)}

The average precision (AP) for a specific IoU threshold T is the area under the precision-recall curve. 
The average recall (AR) for a fixed number of detections per image $N$ is the recall achieved when the model outputs $N$ detections. 
The mAP, and respectively mAR, are then calculated as the mean of the AP, respectively the AR, across a range of IoU thresholds:

\begin{equation}
    \begin{aligned}
        \text{mAP} &:= \frac{1}{M} \sum_{m=1}^{M} \text{AP}(T_m) \\
        \text{mAR} &:= \frac{1}{M} \sum_{n=1}^{M} \text{AR}(T_m) \\
    \end{aligned}
\end{equation}

\noindent where $T_m$ is the m-th IoU threshold and M is the total number of thresholds. 

\begin{figure*}[tbp]
\centering
\includegraphics[width=0.75\textwidth]{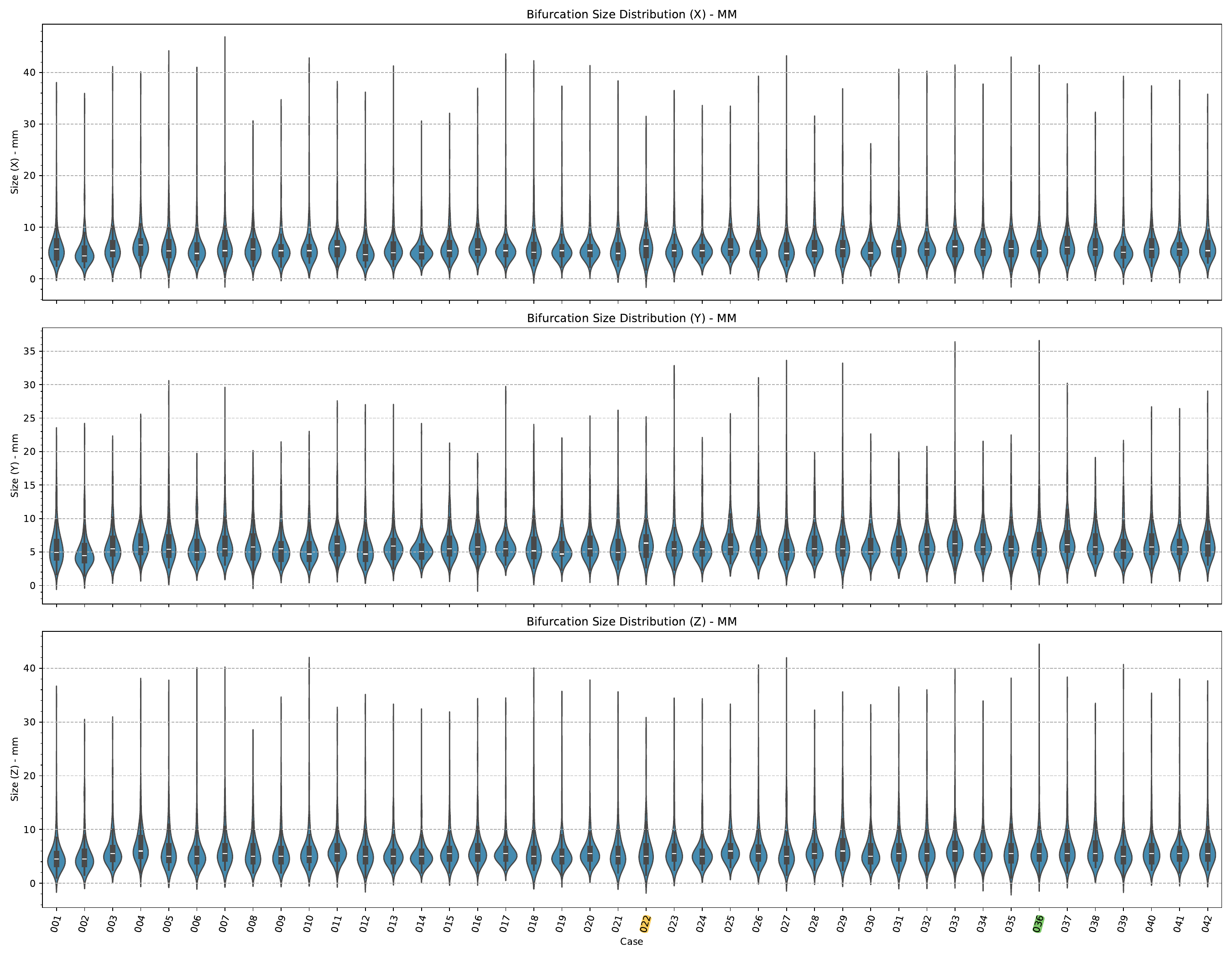}
\label{fig:violin_mm}
\end{figure*}

\end{document}